\RequirePackage{fix-cm}
\documentclass{article}

\usepackage{graphicx}
\usepackage{mathptmx}      
\usepackage{algorithm}
\usepackage{algpseudocode}
\usepackage[colorlinks,linkcolor=black,anchorcolor=black,citecolor=black,urlcolor=blue]{hyperref}
\usepackage{mathtools}
\usepackage{amssymb}
\usepackage{mciteplus}
\usepackage{cite}
\usepackage{soul}
\usepackage{caption}
\usepackage{subcaption}
\usepackage{float}

\usepackage{geometry}
\usepackage[utf8]{inputenc}
\usepackage{algpseudocode}
\usepackage{algorithm}
\usepackage{hyperref}
\usepackage{mathrsfs}

\usepackage{achemso}
\setkeys{acs}{articletitle=true}

\hypersetup{
    colorlinks=true,
    linkcolor=blue,
    filecolor=magenta,      
    urlcolor=magenta,
    citecolor=blue
    }

\title{Graph-Based Bidirectional Transformer Decision Threshold Adjustment Algorithm for Class-Imbalanced Molecular Data}
\author{Nicole Hayes$^1$, 
Ekaterina Merkurjev$^{1,2}$\footnote{Corresponding author,
	Email:  merkurje@msu.edu} ~ and 
 Guo-Wei Wei$^{1,3,4}$\\
$^1$ Department of Mathematics, \\
Michigan State University, MI 48824, USA.\\
$^2$ Department of Computational Mathematics, Science and Engineering\\
Michigan State University, MI 48824, USA.\\
$^3$ Department of Electrical and Computer Engineering,\\
Michigan State University, MI 48824, USA. \\
$^4$ Department of Biochemistry and Molecular Biology,\\
Michigan State University, MI 48824, USA. \\
}

\begin{document}

\maketitle

\abstract{Data sets with imbalanced class sizes, where one class size is much smaller than that of others, occur exceedingly often in many applications, including those with biological foundations, such as disease diagnosis and drug discovery. Therefore, it is extremely important to be able to identify data elements of classes of various sizes, as a failure to do so can result in heavy costs. Nonetheless, many data classification procedures do not perform well on imbalanced data sets as they often fail to detect elements belonging to underrepresented classes. In this work, we propose the BTDT-MBO algorithm, incorporating Merriman-Bence-Osher (MBO) approaches and a bidirectional transformer, as well as distance correlation and decision threshold adjustments, for data classification tasks on highly imbalanced molecular data sets, where the sizes of the classes vary greatly. The proposed technique not only integrates adjustments in the classification threshold for the MBO algorithm in order to help deal with the class imbalance, but also uses a bidirectional transformer procedure based on an attention mechanism for self-supervised learning. In addition, the model implements distance correlation as a weight function for the similarity graph-based framework on which the adjusted MBO algorithm operates. The proposed method is validated using six molecular data sets and compared to other related techniques. The computational experiments show that the proposed technique is superior to competing approaches even in the case of a high class imbalance ratio. \\ \\
{\it Keywords:} Imbalanced data; molecular data; transformer; graph-based; data classification.
}

\section{Introduction}

In data classification, class imbalanced data is distinguished by an underrepresentation of one class (the  ``minority" class) in the data compared to the other class or classes (the  `` majority" class(es)). As a result, data classification algorithms trained on imbalanced data tend to become biased with regard to the majority class(es) and often misclassify data points from the minority group, which is frequently the class of most interest \cite{esposito2021, haixiang2016}. This scenario is quite common in classification tasks, and it can be especially problematic if the minority class represents an important or critical outcome, such as in drug discovery, disease diagnosis, fraud detection, energy management, and other applications involving rare events \cite{haixiang2016}. For example, drug discovery is a very exacting problem in science. In drug screening, the number of inactive molecules is typically thousands of times larger than that of active molecules. A successful drug must be highly potent, be free of serious side effects or off-target binding, and exhibit minimal or no toxicity \cite{esposito2021}. As a result, the majority of drug candidates are either inactive or unqualified, leading to highly imbalanced data sets \cite{lopez2021balancing,guan2022class}. Specifically, imbalanced data occur during virtual screening: identifying potential drug candidates from large compound libraries. Effective handling of imbalanced data can improve the identification of active compounds \cite{korkmaz2020,xu2024computational, tian2022pharmacophore,nair2022deep}. Another situation in which we deal with imbalanced data is quantitative structure-activity relationship modeling, which predicts a compound's activity using the chemical composition \cite{casanova2021novel,gu20233d}. Balancing the data can lead to more reliable predictions \cite{soares2022re}. \\

Therefore, addressing imbalanced data is crucial for effective machine learning in many applications such as drug design and discovery. By employing a fusion of algorithm-level and data-level strategies, and focusing on appropriate evaluation metrics, researchers can build models that better identify rare but valuable compounds, leading to more successful outcomes in drug discovery and development.\\


Techniques to deal with imbalanced data can be categorized into: \cite{haixiang2016, kaur2019, sun2009, ganganwar2012}: data preprocessing or resampling methods\cite{estabrooks2004,garcia2012}, model-level techniques \cite{haixiang2016,fernandez2018}, ensemble algorithms \cite{galar2011}, and cost-sensitive methods \cite{elkan2001}. Strategies in the first group include resampling methods, which attempt to make the training set's class distribution more balanced using sampling techniques, and feature selection methods, which are carried out in the feature space rather than the sample space \cite{haixiang2016}. Examples of resampling procedures include oversampling \cite{cao2013,nekooeimehr2016}, undersampling \cite{kumar2014, anand2010} and hybrid-sampling algorithms \cite{estabrooks2004, song2016,cateni2014}. Respectively, these procedures attempt to rebalance the sample space by creating new samples from the minority group, discarding representatives from the majority group, or a combination of both.
Methods belonging to the second group, the model-level techniques, attempt to make adjustments to steps of the algorithm itself so that there is less bias towards classes of larger size. Ensemble algorithms use the output of multiple classifiers to make predictions, which can improve the performance of individual classifiers \cite{haixiang2016, lopez2013}. Lastly, cost-sensitive procedures instead assign a larger misclassification cost to classes of smaller size \cite{elkan2001}.\\


Some cost-sensitive approaches to imbalanced data classification involve thresholding, which is applied in a postprocessing step \cite{sheng2006}. In particular, thresholding algorithms are cost-sensitive techniques which assign a probability to each of the test elements. For a threshold of 0.5, which is the default probability threshold value in most classifiers \cite{zou2016}, any test element with a probability of at least 0.5 will be assigned to one class; otherwise, it will be assigned to another class. In the case of highly imbalanced data classification, the threshold of 0.5 may not be optimal; thus, threshold algorithms attempt to find a more optimal threshold boundary for the machine learning task. Usually, a balanced accuracy metric is used; some examples include the ROC curve (receiver operating characteristic curve) \cite{fawcett2006}, the G-Mean \cite{johnson2021}, F1-score \cite{zou2016}, the Matthews correlation coefficient \cite{esposito2020} and the balanced accuracy \cite{korkmaz2020}. In these cases, the goal is to maximize the balanced accuracy metric. Overall, one approach for image classification and feature extraction is detailed in \cite{yin2024conv}, which proposes a convolution-based efficient transformer image feature extraction network. Moreover, Wang et al.  \cite{wang2024stacked} introduce a micro-directional propagation model using deep learning, and Zhu et al. \cite{zhu2023adaptive} focus on an adaptive agent decision model derived using autonomous learning deep reinforcement learning. In addition, Ref. \cite{yinafbnet} details feature extraction modules, while Ref. \cite{peng2023} presents a systematic review of picture fuzzy decision-making methods. Applications to sideband harmonics, internet market design, large-scale sliding puzzles are found in \cite{jiao2024}, \cite{cheng2023truthfulness}, and \cite{wang2017}. \\ 

However, most of the cost-sensitive techniques for imbalanced data classification that search for the optimal threshold can be computationally extensive because retraining of the classifier is often required. Inspired by the work of Esposito et al. \cite{esposito2021}, in this paper, we detail an algorithm for imbalanced data classification which does not require any retraining in its procedure. In order to motivate the proposed procedure, we turn to the authors' previous work \cite{hayes2023}, which demonstrated the success of their BT-MBO method for molecular data classification with very small labeled sets; refer to \cite{dou2023} for a survey of machine learning techniques for data with limited labeled elements in molecular science. The BT-MBO algorithm, which admits two dimensional molecular data in a special format called the simplified molecular input line entry specification (SMILES) \cite{smiles} format, is composed of a bidirectional transformer \cite{hundreds} and a graph-based and modified Merriman-Bence-Osher (MBO) algorithm \cite{mbo,mbo1994} in series, utilizing the transformer model to generate molecular fingerprints which then serve as features for the MBO algorithm. The MBO method used in the BT-MBO model \cite{hayes2023} generates a probability distribution over all classes for each data point, and it subsequently classifies each data element by choosing the class for which that element has the highest probability. However, in the case of imbalanced data, this procedure may cause the technique to miscategorize data elements in the minority group, since the probability distribution may be biased toward the majority classes. Thus, in the present work, we propose the BTDT-MBO method (bidirectional transformer with MBO techniques using a varying decision threshold), which is a novel adaptation of the BT-MBO technique for use on highly imbalanced molecular data.\\

In particular, the proposed BTDT-MBO method incorporates a decision threshold adjustment in the MBO algorithm, which utilizes the probability distribution generated by the algorithm for each data point. Instead of following the typical MBO classification procedure, the BTDT-MBO algorithm tests a set of thresholds and chooses the threshold that produces the highest ROC-AUC score (area under the ROC curve) for the unlabeled points. The new thresholding step introduced in the BTDT-MBO method does not require any additional iterations of the MBO algorithm; rather, at the end of each iteration, we repeat the new thresholding step for each threshold. We refer to the modified MBO algorithm with decision threshold adjustment as the DT-MBO method. Another adaptation introduced in the BTDT-MBO method is the inclusion of the distance correlation \cite{szekely2007, ccp} as a weight function in the DT-MBO algorithm. The default weight function used in our BTDT-MBO method is a Gaussian kernel; in addition to experiments using a Gaussian weight function, we also carried out experiments using a distance correlation weight function. \\ 

Given our previous BT-MBO method's predictive success for the challenging scenario of scarcely labeled data \cite{hayes2023}, we focus on the difficult task of data classification of highly imbalanced molecular data sets in the present paper. Specifically, all six data sets used for benchmarking the proposed BTDT-MBO algorithm have an imbalance ratio (IR) of 16.5 or greater. We compare our results using BTDT-MBO to those of the GHOST procedure \cite{esposito2021} applied to four common classification algorithms: random forest, gradient boosting decision trees, extreme gradient boosting, and logistic regression. Specifically, the GHOST technique \cite{esposito2021} is an automated procedure that can be used in any data classification procedure and that adjusts the decision threshold of a classifier using the training set's prediction probabilities (i.e., without needing to retrain the classifier); it achieved success in classifying imbalanced molecular data, demonstrated on data with varying levels of class imbalance. Detailed results on the six highly imbalanced data sets, as well as a discussion, are given in Section \ref{results}. The computational experiments indicate that our BTDT-MBO method obtains superior results to those of the comparison techniques. \\

\subsection*{Contributions} 

We present the main contributions of this work:

\begin{itemize}

\item We present a new technique, the BTDT-MBO method, for molecular data classification, especially designed to be very useful even in cases of highly imbalanced data sets. This procedure incorporates Merriman-Bence-Osher techniques, as well as bidirectional transformers, distance correlation, and adjustments in the classification threshold.

\item The proposed BTDT-MBO algorithm is able to perform extremely well even in the cases of molecular data that is highly imbalanced and where the class sizes vary significantly. This is extremely important, since highly imbalanced data sets are often found in practice.

\item The proposed method utilizes several techniques to achieve good performance when using data with high class imbalance ratios, including adjustments in the classification threshold. 

\item The proposed technique is evaluated using experiments on six data sets. It is shown that the BTDT-MBO method performs more accurately than the comparison methods in almost all cases. The ROC-AUC score is used as a metric. 

\end{itemize}

\section{Methods}\label{methods}

In this section, we derive the proposed techniques. In particular, in Section \ref{btdt-mbo}, we outline the details of the proposed BTDT-MBO method (bidirectional transformer with MBO techniques \cite{mbo,mbo1994} using a varying decision threshold), which adapts some of the techniques of our previous work \cite{hayes2023}. In Section \ref{distcorr}, we further discuss the distance correlation \cite{ccp, szekely2007} as a potential similarity metric for use in the BTDT-MBO method.\\

\subsection{BTDT-MBO Algorithm}\label{btdt-mbo}

The authors' previous work \cite{hayes2023} demonstrated the success of their proposed BT-MBO method in molecular classification problems with small sets of labeled nodes, with the BT-MBO procedure outperforming state-of-the-art techniques on some benchmark data sets when only one percent of all data elements are viewed as labeled. The present work proposes a method, called BTDT-MBO, that adapts some of the procedures of the BT-MBO procedure for the data classification task on molecular data sets where the classes of the data set vary vastly in size. \\

\begin{figure}
    \centering
    \hspace{-0.2cm}
    \includegraphics[width=12.75cm]{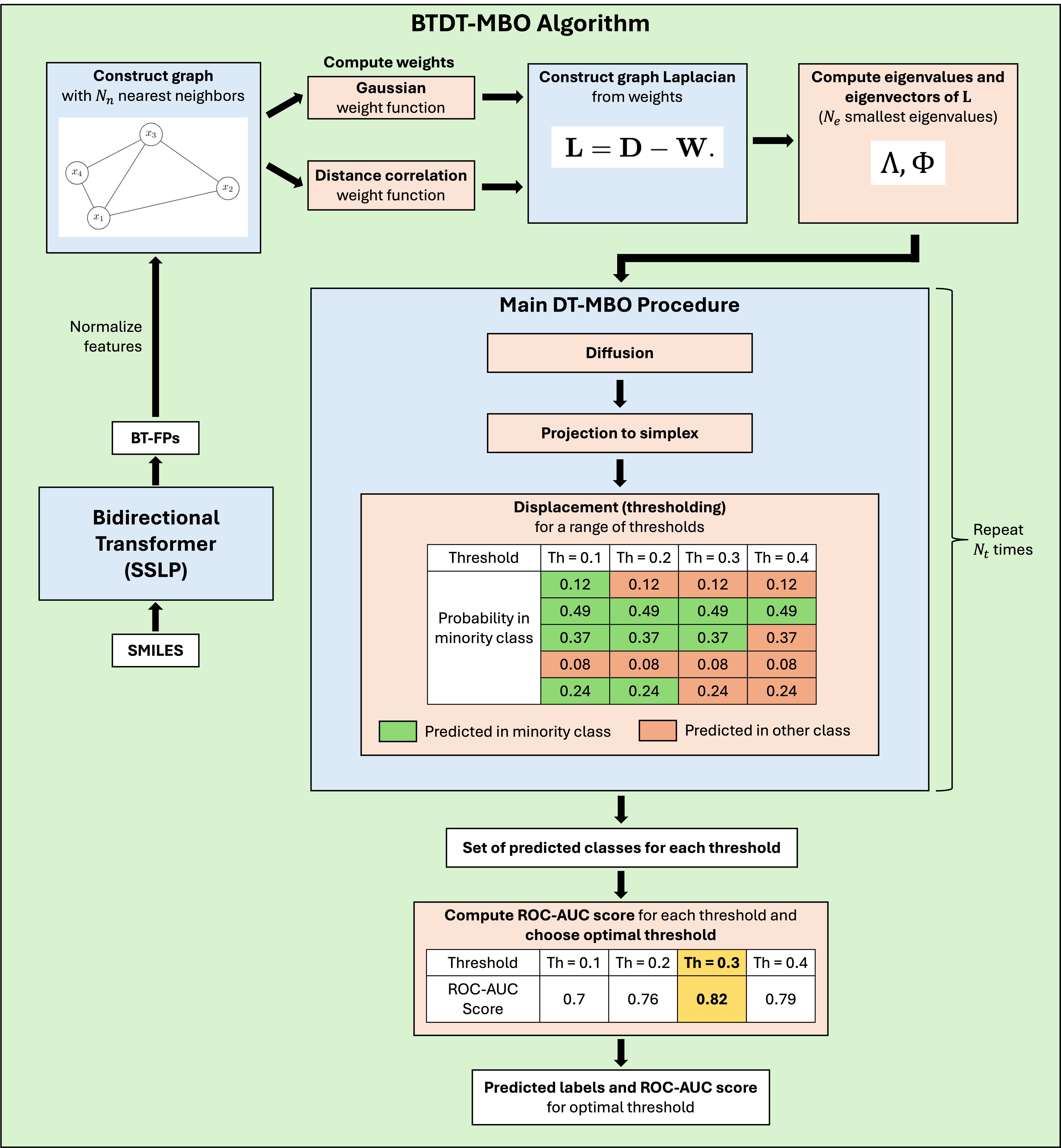}
    \caption{Flowchart illustrating the BTDT-MBO method.}
    \label{flowchart}
\end{figure}

Inspired in part by the success of Esposito et al. \cite{esposito2021} in adjusting the classification threshold for machine learning classifiers to make predictions about imbalanced molecular data, the proposed BTDT-MBO method utilizes the aspects of the BT-MBO algorithm from \cite{hayes2023} along with a decision threshold adjustment at each iteration of the method. We denote the adjusted MBO method with varied decision threshold by DT-MBO. The proposed BTDT-MBO algorithm consists of a bidirectional transformer (created by Chen et al. \cite{hundreds}) and a DT-MBO algorithm in series, where the fingerprints (BT-FPs) generated by the transformer serve as input features to the DT-MBO algorithm. Figure \ref{flowchart} illustrates the steps of the BTDT-MBO algorithm. The BTDT-MBO algorithm is further outlined in Algorithm \ref{alg-bt}, and details of the DT-MBO method are given in Algorithm \ref{alg-dtmbo}. \\

\subsubsection{MBO Scheme with Decision Threshold Adjustment}\label{mbo}

Here, we establish the similarity graph-based structure our BTDT-MBO scheme uses, which adapts techniques outlined in the literature \cite{kostic, merkurjev2021multiscale, mbo}. First, consider a graph involving the vertex set $V$ representing the data set nodes. We can then define a weight function $w:V\times V \to \mathbb{R}$, whose value represents the degree of similarity between two vertices of the graph. Specifically, $w(i,j)$ measures the degree of similarity of data elements $i$ and $j$ of the data set. While there are many choices for the weight function, it should be constructed so that the value of $w(i,j)$ corresponds to the extent of resemblance or closeness of the data elements $i$ and $j$. A popular weight function is the Gaussian weight function: 
\begin{equation}
\label{gaussian}
    w(i,j) = \text{exp} \left( - \frac{d(i,j)^2}{\sigma^2} \right),
\end{equation}
where $d(i,j)$ represents the distance between vertices $i$ and $j$ (computed using some prespecified metric such as the Euclidean distance), and $\sigma>0$ \cite{mbo, merkurjev2021multiscale}.\\

In this paper, we utilize the Gaussian weight function for some of our BTDT-MBO experiments, and for other experiments, we use the distance correlation \cite{szekely2007} as a weight function. The distance correlation is defined and further discussed in Section \ref{distcorr}.\\

Given a weight function $w$, let the weight matrix $\mathbf{W}$ be the matrix $\mathbf{W}_{ij} = w(i,j)$. Then let the degree of a vertex $i \in V$ be defined as $d_i = \sum_{j \in V} w(i,j)$. If we let $\mathbf{D}$ be a diagonal matrix containing diagonal elements $\{d_i\}$, then we can define the graph Laplacian as

\begin{equation}
\label{laplacian}
    \mathbf{L} = \mathbf{D} - \mathbf{W}.
\end{equation}

Occasionally, the graph Laplacian is normalized so that the diagonal entries of the normalized Laplacian are unit values. Of course, the non-diagonal entries are scaled accordingly. \\

For outlining the Merriman-Bence-Osher (MBO) framework, formerly introduced in \cite{mbo1994} to generate approximations to mean curvature flow and then applied using a graph-based setting to various applications such as hyperspectral imaging in works such as \cite{kostic, mbo, meng2017,merkurjev2018,merkurjevhyperspectral}, we consider a general data classification problem with $m$ classes. Overall, the MBO algorithm ultimately aims to compute a label matrix $\mathbf{U} = (\mathbf{u}_1, \dots , \mathbf{u}_N)^T$; a row $\mathbf{u}_i \in \mathbb{R}^m$ of the label matrix contains the probability values of the data element $i$ belonging to each of the $m$ classes. For instance, the first element of $\mathbf{u}_i$ represents the probability that node $i$ is a member of the first class, the second element of $\mathbf{u}_i$ represents the probability that node $i$ is a member of the second class, and so on. Therefore, each vector $\mathbf{u}_i$ can be viewed as a component of the $m^{th}$ Gibbs simplex:

\vspace{-0.2cm}
\begin{equation}
    \label{gibbs}
    \Sigma^m := \{(y_1, ..., y_m) \in [0,1]^m \; \;\text{such that} \; \; \sum_{k=1}^m y_k = 1\}
    \vspace{0.2cm}
\end{equation}

The vertices of $m^{th}$ Gibbs simplex are unit vectors $\{\mathbf{e}_i\}$, where the probability of belonging to class $i$ is 1, while the probability of belonging to any other class is 0. Thus, the vertices $\{\mathbf{e}_i\}$ correspond to data elements that belong exclusively to each of the $m$ classes. In our data classification problem with labeled and unlabeled data elements, the labeled data elements are assigned to their corresponding vertices of the Gibbs simplex.\\ 

The process by which the MBO algorithm generates the optimal label matrix $\mathbf{U}$ relies on energy minimization, a time-splitting scheme, and transference to a graph-based setting as described above. A general graph-based machine learning algorithm for data classification can be formulated as minimizing the energy consisting of the sum of a regularization term that incorporates weights in addition to a fidelity part that integrates the labeled nodes \cite{merkurjev2021multiscale}.\\

As Garcia and coauthors demonstrated in \cite{mbo}, one successful choice for the regularization term $\text{R}$ is the Ginzburg-Landau functional \cite{ginzburg1,ginzburg2}, often utilized in image processing tasks and associated applications due to its relation to the total variation functional \cite{kostic, tsai}. For the fidelity term, one can choose an $L_2$ fit to all labeled nodes. In the continuous setting, $L_2$ gradient descent can then be applied to minimize the energy. This results in a modified Allen-Cahn equation containing an additional forcing part. Next, a time-splitting scheme produces a technique that alternates between two steps: a diffusion step, which uses the heat equation with an addition term, and a thresholding step. Finally, one can use the techniques outlined in \cite{mbo, kostic, merkurjev_aml} to transfer the procedure to a graph-based setting, in which the thresholding step is replaced by projection to the Gibbs simplex (\ref{gibbs}), followed by displacing to the nearest vertex in the $m^{th}$ Gibbs simplex. Additionally, to adapt to the graph setting, one can replace the Laplace operator by the graph Laplacian (\ref{laplacian}) or some normalization of the graph Laplacian. \\

  Thus, each iteration of the MBO algorithm ends with the aforementioned displacement step, where each row of the label matrix is displaced to its closest vertex in the simplex. In the algorithm, the closest vertex is determined by the class for which the data element corresponding to that row has the highest probability. In this paper, we propose that this procedure can instead be formulated using a decision threshold, which utilizes the probability that a data element belongs to the minority class. Given a decision threshold and a row of the label matrix, if the probability corresponding to the minority class is greater than that threshold, that row will be displaced to the vertex for the minority class. Otherwise, the typical displacement procedure is followed, with the row being displaced to the vertex for which it has the highest probability.\\

To formulate our decision threshold technique, we are inspired by Esposito et al. \cite{esposito2021}, which demonstrated success in classifying imbalanced molecular data by developing a procedure that adjusts and optimizes the decision threshold of machine learning classifiers. For binary data, the default decision threshold is usually 0.5, where the model assigns a sample to the minority class if its probability of belonging to the minority class is greater than 0.5 (this is identical to assigning the sample to the class for which it has the highest probability). However, machine learning classifiers are likely to misclassify minority points since the probability distribution over the two classes for a given data point may be skewed toward the majority class \cite{king2002}, so Esposito et al. \cite{esposito2021} tested potential decision thresholds from 0.05 to 0.5 with a spacing of 0.05. Similarly, in the present work, we augment the MBO method by a process that optimizes the decision threshold for the displacement step, resulting in our proposed decision-threshold-MBO (DT-MBO) algorithm.\\

Two of the inputs of the DT-MBO algorithm, which is integrated into the proposed BTDT-MBO procedure, are a minimum threshold and a maximum threshold to test. In the present study, we test a minimum threshold of 0.05 and a maximum threshold of 0.55, with thresholds between these two values spaced by 0.05. This allows us to choose a more optimal threshold than the usual one of $0.5$. \\

Overall, we summarize the main iterative steps of the DT-MBO procedure for the data classification task used in the present work below, with more details given in Algorithm \ref{alg-dtmbo}. The process is adapted from the MBO algorithm used in our previous work \cite{hayes2023}, with an additional decision threshold step in the displacement step. The below steps require a given labeled/unlabeled partition of a data set as well as a list of thresholds to test.

\begin{enumerate}
    \item Diffusion step: compute $\mathbf{U}^{n+\frac{1}{2}}$ using the heat equation with an additional term using $$\mathbf{U}^{n+\frac{1}{2}} = \mathbf{U}^{n} - \text{dt}{\mathbf{L}\mathbf{U}^{n+\frac{1}{2}}} + \mu \cdot (\mathbf{U}^{n} - \mathbf{U}_{\text{labeled}}),$$ applied $N_s$ times. Here, $\text{dt}>0$, and $\mathbf{U}_{\text{labeled}}$ is a matrix where a row corresponding to a labeled element is an indicator vector $\mathbf{e}_k$ if the node is of class $k$, and a zero vector for non-labeled points. Moreover, $\mathbf{\mu}$ is a vector whose entries are equal to a constant for labeled points, and zero otherwise. We set $N_s=3$.
    \vspace{0.1cm}
    \item Projection step: compute $\mathbf{U}^{n+1}$ via projecting (each row) of the matrix $\mathbf{U}^{n+\frac{1}{2}}$ onto (\ref{gibbs}).
        \vspace{0.1cm}
    \item Displacement step: Replace every row of the result from step 2 by its assigned vertex $\mathbf{e}_k$ in the $m^{th}$ Gibbs simplex (\ref{gibbs}) depending on the decision threshold. This procedure is executed for every threshold in the list.
            \vspace{0.1cm}
    \begin{enumerate}
        \item If the probability that the row belongs to the minority class is greater than (or equal to) the decision threshold, the row is replaced by the vertex corresponding to the minority class.
                \vspace{0.1cm}
        \item Otherwise, the row is replaced by the vertex for which it has the highest probability.
                \vspace{0.1cm}
    \end{enumerate} 
\end{enumerate}

After repeating the above steps for a number of iterations, our model generates a predicted label matrix for each threshold. To select the optimal decision threshold from the set of thresholds for a given labeled/unlabeled partition of a data set, our model then computes the ROC-AUC score for each label matrix and subsequently chooses the threshold whose predicted labels yield the highest ROC-AUC score. Notably, the new thresholding procedure only involves alterations to the displacement step and does not require additional iterations of the diffusion or projection to simplex steps, thus saving computational time, in contrast to many cost-sensitive methods which require the complete retraining of the classifier. Section \ref{performance-and-discussion} contains discussion regarding this threshold optimization and its results.  

 \subsubsection{Bidirectional Transformer}\label{bt}

As in our previous work \cite{hayes2023}, the BTDT-MBO algorithm utilizes a bidirectional transformer to convert the SMILES input for each molecular compound to a vector in a latent space, which is then obtained as a molecular fingerprint, named BT-FP, for that compound. After being normalized, the BT-FPs for the compounds in a given data set comprise the features passed to the DT-MBO method for that data set. Specifically, the BTDT-MBO algorithm uses a particular bidirectional transformer model introduced by Chen et al. \cite{hundreds} involving an attention structure for self-supervised learning (SSL). The model, the self-supervised learning platform (SSLP), enables the generation of BT-FPs without the need for data labels. Moreover, the model used for BTDT-MBO can be chosen out of three SSLP models constructed by Chen et al. \cite{hundreds}. The three SSLP models include one pretrained on the ChEMBL data set \cite{chembl-source}. The other two models include one pretrained on the union of the ChEMBL and PubMed \cite{pubchem-source} data sets and one model trained on the union of the ChEMBL, PubMed, and ZINC \cite{zinc-source} data sets. Additionally, there is an option to fine-tune the selected self-supervised learning platform for specific predictive tasks.\\

In our prior work \cite{hayes2023}, we demonstrated the predictive success of the BT-MBO model utilizing the self-supervised learning platform trained solely on ChEMBL for scarcely labeled molecular data. Consequently, the proposed BTDT-MBO model in the present work uses the same SSLP trained on ChEMBL. Similarly, our BT-MBO model in \cite{hayes2023} performed well without fine-tuning the SSLP, so we also bypass the fine-tuning step for the proposed BTDT-MBO model, instead passing our data directly to the pretrained SSLP.  

\begin{algorithm}[H]
\caption{DT-MBO Algorithm (using techniques from \cite{merkurjev_aml, merkurjev2021multiscale})}\label{alg-dtmbo}
\begin{algorithmic}[H]
\vspace{0.1cm}
\Require  Labeled set $\mathscr{L} = \{(\mathbf{x}_i, y_i)\}_{i=1}^N$, with fingerprint $\mathbf{x}_i$ and label $y_i$, $N$= size of data set, $N_n=$number of nearest neighbors, $N_e=$number of eigenvectors, $dt$, $C$, $N_t=$ maximum number of iterations, $N_p=$ number of labeled elements, $N_l=$ number of labeled collections, $N_s$, $m=$ number of classes, $\tau_{\text{low}}=$ lowest threshold, $\tau_{\text{high}}=$ highest threshold, $k_\text{min}=$ index of minority group/cluster.
 \vspace{0.1cm}
\Ensure  Average optimal ROC-AUC score over the $N_l$ labeled sets.
\vspace{0.15cm}
\State 1: Compute the $N_n$-nearest neighbor graph, unless one is using the Nystr\"{o}m technique \cite{nystrom1,nystrom2,nystrom3} for large data. In the latter case, go to Step 3.
\vspace{0.15cm}
\State 2: Construct the graph Laplacian $\mathbf{L}$.
\vspace{0.15cm}
\State 3: Calculate the smallest $N_e$ eigenvalues $\Lambda$ and the corresponding eigenvectors $\Phi$ of $\mathbf{L}$. One can also approximate them using the Nystr\"{o}m method \cite{nystrom1,nystrom2,nystrom3} in case of large data sets. 
\vspace{0.15cm}
\State 4: The main procedure of the proposed method:
\vspace{0.15cm}
\For{$l = 1 \to N_l$}
\vspace{0.1cm}
\State Let the $N \times 1$ vector $\Gamma$ vector be defined so that $\Gamma_j = 1$ for labeled points and $0$ for unlabeled points.
\vspace{0.125cm}
\For{$i = 1 \to N$}
\vspace{0.125cm}
\State Initialize each value of a row of $\mathbf{U}$ to a random number in [0,1], except for rows corresponding to labeled nodes, in which case, set $\mathbf{u}_i^0 \leftarrow \boldsymbol{e}_k$, where $k$ is the true class.
\vspace{0.125cm}
\State $\mathbf{u}_i^0 \leftarrow projectToSimplex(\mathbf{u}_i^0)$. Here, $\mathbf{u}_i$ is $i^{th}$ row of the matrix $\mathbf{U}^0$.
\vspace{0.1cm}
\EndFor
\vspace{0.125cm}
\State $n_\tau \leftarrow (\tau_{\text{high}} - \tau_{\text{low}})/0.05 + 1$, \hspace{0.2cm}  $\mathbf{U}_\tau \leftarrow \mathbf{0}$, an $N \times m \times n_\tau$ array of zeros.
\vspace{0.1cm}
\State $\mathbf{A} \leftarrow \Phi'\mathbf{U}^0$,  \hspace{0.2cm} $\mathbf{B} \leftarrow \mathbf{0}$, \hspace{0.2cm}  $\mathbf{E} \leftarrow 1 + (dt/N_s)\Lambda$
\vspace{0.2cm}
\For{$n = 1 \to N_t$}
\vspace{0.1cm}
\For{$k = 1 \to N_s$}
\vspace{0.1cm}
\For{$j = 1 \to m$}
\vspace{0.1cm}
\State $\mathbf{A}_j \leftarrow (\mathbf{A}_j- (dt/N_s)\mathbf{B}_j)./\mathbf{E}$
\vspace{0.1cm}
\EndFor
\vspace{0.1cm}
\State $\mathbf{U}^n \leftarrow \Phi \mathbf{A}$
\vspace{0.1cm}
\State $\mathbf{B} \leftarrow \mathbf{C} (\Phi' (\Gamma \cdot (\mathbf{U}^n - \mathbf{U}^0))) $, where row-wise multiplication is performed.
\vspace{0.1cm}
\EndFor
\vspace{0.1cm}
\State $\mathbf{U}^{n+1} \leftarrow projectToSimplex(\mathbf{U}^{n})$
\vspace{0.1cm}
\For{$j = 1\to n_\tau$}
\vspace{0.1cm}
\State $\tau_j = \tau_{\text{low}} + 0.05(j-1)$
\vspace{0.1cm}
\For{$i = 1 \to N$}
\vspace{0.1cm}
\If{$\mathbf{u}_{ik_\text{min}}^{n+1} \geq \tau_j$}
\State $\mathbf{u}_i^{n+1} \leftarrow \mathbf{e}_{k_\text{min}}$, where $\mathbf{e}_{k_\text{min}}$ is the simplex vertex corresponding to the minority class.
\Else
\State $\mathbf{u}_i^{n+1} \leftarrow \mathbf{e}_k$, where $\mathbf{e}_k$ is the closest simplex vertex to $\mathbf{u}_i^{n+1}$.
\EndIf
\vspace{0.1cm}
\EndFor
\vspace{0.1cm}
\State $\mathbf{U}_{\tau_j} \leftarrow \mathbf{U}^{n+1}$, where $\mathbf{U}_{\tau_j}$ is the $j$th $N \times m$ slice of $\mathbf{U}_{\tau_j}$.
\vspace{0.1cm}
\EndFor
\vspace{0.1cm}
\EndFor
\vspace{0.1cm}
\State Compute ROC-AUC score for each threshold, and choose the threshold that yields the highest score.
\vspace{0.1cm}
\EndFor
\vspace{0.1cm}
\State 5: Compute the average of the $N_l$ optimal ROC-AUC scores.
\vspace{0.1cm}
\end{algorithmic}
\end{algorithm}

\begin{algorithm}[h!]
\caption{BTDT-MBO Method (using related procedures as in \cite{hayes2023})}\label{alg-bt}
\begin{algorithmic}[H]
\vspace{0.125cm}
\Require  Unlabeled set $\mathscr{U} = \{s_i\}_{i=1}^N$, where $s_i$ is a molecular compound's SMILES string), a chosen pre-trained model from the three described in Section 2.1.2, a dictionary which designates an integer number to each SMILES element.
\vspace{0.125cm}
\Ensure  $\mathscr{F}_{\text{BT}} = \{\mathbf{x}_i\}_{i=1}^N$, a set of normalized fingerprints.
\vspace{0.125cm}
\State 1: Binarize the data consisting of SMILES strings.
\vspace{0.125cm}
\State 2: Loading step: load the chosen model (which is pre-trained).
\vspace{0.125cm}
\State 3: Extracting step: Extract the hidden information from $\mathscr{U}$ via the chosen pretrained model:
\vspace{0.125cm}
\State Initialize the empty dictionary $D_\text{hidden}$ (of length $N$) containing hidden features.
\vspace{0.125cm}
\For{$i = 1 \to N$}
\vspace{0.125cm}
\State Compute tensor $\mathbf{t}_i$ (of the same dimension as $s_i$), containing the dictionary values of the characters in the analogous locations in $s_i$.
\vspace{0.125cm}
\State Let $\mathbf{t}_i$ be an input to the bidirectional encoder transformer from the chosen pretrained model. Obtain the inner state from the last hidden layer $\mathbf{T}_i$, whose dimensions are $(l_i, 1, 512)$ (where $l_i$ is the length of $\mathbf{t}_i$).
\vspace{0.125cm}
\State Reshape $\mathbf{T}_i$ into a tensor $\mathbf{V}_i$ with the following dimensions: $(l_i, 512)$.
\vspace{0.125cm}
\State Set $\mathbf{V}_i$ as $i$th element of the matrix $D_\text{hidden}$.
\vspace{0.125cm}
\EndFor
\vspace{0.125cm}
\State 4: Generating step: Produce the fingerprints from the hidden information:
\vspace{0.125cm}
\State Initialize the list of fingerprints $\mathscr{F}_{\text{BT}}$ (of length $N$).
\vspace{0.1cm}
\For{$i = 1 \to N$}
\vspace{0.125cm}
\State Compute the first row $\mathbf{x}_i$ from the $i$th element of the matrix $D_\text{hidden}$.
\vspace{0.125cm}
\State Set$\mathbf{x}_i$ as the $i$th fingerprint in $\mathscr{F}_{\text{BT}}$.
\vspace{0.125cm}
\EndFor
\vspace{0.125cm}
\State 5: Scale fingerprints $\mathscr{F}_{\text{BT}} = \{\mathbf{x}_i\}_{i=1}^N$ so that they have unit variance and zero mean.
\vspace{0.125cm}
\State 6: Send scaled fingerprints $\mathscr{F}_{\text{BT}} = \{\mathbf{x}_i\}_{i=1}^N$ to the DT-MBO algorithm, described in Algorithm \ref{alg-dtmbo}.
\vspace{0.125cm}
\end{algorithmic}
\end{algorithm}

\subsubsection{Overall Procedure} 

The overall two stages of the BTDT-MBO procedure are outlined as follows: 

\begin{itemize}
    \item Molecular data in the form of SMILES strings are passed to the self-supervised learning platform pretrained on ChEMBL data set \cite{chembl-source}.
    \item The resulting BT-FPs are the rescaled so that they have zero mean as well as unit variance and then served as an input to the DT-MBO method, outlined in Algorithm \ref{alg-dtmbo}. 
\end{itemize} 

More details of the stages are given in Algorithm \ref{alg-bt}. Notably, because the SSLP does not require data labels, this procedure can be applied to data with any number of classes. Paired with our DT-MBO method as outlined in Section \ref{mbo} and Algorithm \ref{alg-dtmbo}, our BTDT-MBO model can thus be used for problems with any number of clusters or classes. \\

\subsection{Distance Correlation}\label{distcorr}

As discussed in Section \ref{mbo}, the similarity graph-based framework on which the DT-MBO scheme operates requires a weight function to compute the degree of similarity between two data elements. Various weight functions can be used, but a suitable weight function is defined so that a high degree of similarity between two elements is reflected by a large weight function output, and a low degree of similarity is reflected by a small weight function output \cite{luxberg}. \\

As in the authors' previous work \cite{hayes2023} using the BT-MBO method, our proposed BTDT-MBO method utilizes a Gaussian kernel (\ref{gaussian}) as one of the ways to compute weights for the MBO scheme. By construction, when the distance between two data elements is 0, the Gaussian weight function evaluated at those two data elements equals 1. As the distance between two data elements approaches infinity, the Gaussian weight function approaches 0. In other words, the Gaussian weight function outputs values close to 0 for dissimilar data elements and 1 for identical data elements, with a weight function output closer to 1 reflecting greater similarity. \\

In the present work, we further propose distance correlation \cite{szekely2007} as a potential weight function for the DT-MBO scheme, motivated in part by the success of Hozumi et al. \cite{ccp} in applying distance correlation to feature clustering. Also, as will be illustrated below, the distance correlation weight function scales similarly to the Gaussian weight function, with potential outputs between 0 and 1, and higher weight function values corresponding to greater similarity, supporting our substitution. We hope to provide a new interpretation of the MBO method by incorporating the distance correlation as a weight function.\\

Distance correlation \cite{szekely2007} is a measure of dependence between random vectors that generalizes of the idea of correlation. In particular, distance correlation is defined for pairs of vectors in arbitrary dimension, and the distance correlation of two vectors is 0 if and only if the two vectors are independent. We use the distance correlation between two vectors as defined in \cite{szekely2007} and use the same notation below as in \cite{ccp}. First, given a vector $\mathbf{z}^i$, $i = 1, 2, \dots I$, we can compute a distance matrix with entries defined by

\begin{equation}
    a^i_{jk} = ||z^i_m - z^i_k||, \quad m, k = 1, 2, \dots M.
\end{equation}

Here, $||\cdot||$ represents the Euclidean norm. Also, the doubly centered distance for $\mathbf{z}^i$ is given by

\begin{equation}
    A^i_{jk} := a_{jk} - \bar{a}_{j\cdot} - \bar{a}_{\cdot k} + \bar{a}_{\cdot \cdot},
\end{equation}

where $\bar{a}_{j\cdot}$ represents the $j$th row mean. In addition, $\bar{a}_{\cdot k}$ represents the $k$th column mean, and $\bar{a}_{\cdot \cdot}$ represents the grand mean of the distance matrix for $\mathbf{z}^i$. \\

Given $\mathbf{z}^i$ and $\mathbf{z}^j$, we can define the squared distance covariance as

\begin{equation}
    \text{dCov}^2(\mathbf{z}^i, \mathbf{z}^j) := \frac{1}{M^2} \sum_j \sum_k A^i_{jk} A^j_{jk} .
\end{equation}

Now, the distance correlation between $\mathbf{z}^i$ and $\mathbf{z}^j$ is defined as

\begin{equation}
    \label{dcor-eqn}
    \text{dCor}(\mathbf{z}^i, \mathbf{z}^j) := \frac{\text{dCov}^2(\mathbf{z}^i, \mathbf{z}^j)}{\text{dCov}(\mathbf{z}^i, \mathbf{z}^i)\text{dCov}(\mathbf{z}^j, \mathbf{z}^j)} .
\end{equation}

Note that the distance correlation $\text{dCor}(\mathbf{z}^i, \mathbf{z}^j)$ can take on values in the range $[0,1]$ (see a complete proof of this fact in \cite{szekely2007}). We have $\text{dCor}(\mathbf{z}^i, \mathbf{z}^j) = 0$ when vectors $\mathbf{z}^i$ and $\mathbf{z}^j$ are independent (or $\text{dCov}^2(\mathbf{z}^i, \mathbf{z}^j) = 0$),  and we have $\text{dCor}(\mathbf{z}^i, \mathbf{z}^j) = 1$ when $\mathbf{z}^i$ and $\mathbf{z}^j$ are linearly dependent. Additionally, a higher squared distance covariance between two vectors corresponds to a distance correlation closer to 1 for those vectors.\\

Thus, the distance correlation (\ref{dcor-eqn}) satisfies the conditions for a similarity weight function for the MBO scheme. As discussed above, the range of the distance correlation is identical to that of the Gaussian kernel, so computationally implementing the distance correlation as a weight function in our DT-MBO method is straightforward. Specifically, only the construction of the graph Laplacian $\mathbf{L}$ in Step 2 of Algorithm \ref{alg-dtmbo} differs when using the distance correlation to compute the graph weights. Furthermore, Sz\'ekely et al. \cite{szekely2007} proved relationships between the distance correlation and the standard bivariate normal distribution which support this substitution in our algorithm. Namely, if $X$ and $Y$ have standard normal distributions with correlation $\rho$, then $\text{dCor}(X, Y) \leq |\rho| $. \\

\section{Results and Discussion}\label{results}

In this section, we discuss the computational experiments and results. The code used for the DT-MBO algorithm was written in MATLAB. 
To carry out computational experiments and optimize parameters, we utilized 
the High Performance Computing Center (HPCC) at Michigan State University. For all jobs submitted through HPCC, we specified two CPUs per task. Distance correlation weights were computed using the \texttt{distance\_correlation} function from the dcor library in Python.

\subsection{Data Sets}\label{data-sets}

The data sets used in the present work are characterized by a high level of imbalance in class sizes, measured by the imbalance ratio (IR). The IR of a data set with two classes is defined as the ratio of the number of points in the majority class to the number of points in the minority class \cite{esposito2021}. Out of the public data sets Esposito et al. \cite{esposito2021} used for benchmarking their proposed thresholding models, we selected data sets from each of the following classes to benchmark our methods:
\begin{itemize}
    \item DS1 Data Sets (obtained from \cite{esposito2021}, introduced in \cite{riniker2013} and refined in \cite{riniker20132}): Each of the DS1 data sets corresponds to a particular ChEMBL target and contains 100 associated diverse active compounds against that target. For each data set, we randomly chose 2,000 inactives from a set of 10,000 assumed inactives drawn from ZINC \cite{zinc1, zinc2}, reproducing the data set construction procedure from \cite{esposito2021}. These 10,000 compounds were selected in \cite{riniker20132} to have similar property distributions to the set of actives, with similarity measured by an atom-count fingerprint. Every DS1 data set has an IR of 20.0 by construction.  
    \item DrugMatrix Assays (obtained from \cite{esposito2021}): Each the 44 DrugMatrix assays benchmarked in \cite{esposito2021} contains results for 842 compounds tested in a particular assay, with each data set labeled by the assay tested. Each molecule was labeled as active or inactive based on its "activity\_comment" value in the source data. In this paper, we evaluate our proposed procedure on four of these data sets, with IRs ranging from 16.5 to 20.0. The source data was retrieved from ChEMBL and was originally recorded in the DrugMatrix database \cite{DrugMatrix}.
\end{itemize}

A brief summary of the data sets used for testing the BTDT-MBO model is given in Table \ref{data-tab}.\\

The authors' BT-MBO method from \cite{hayes2023} performed better than state-of-the-art techniques for scarcely labeled molecular data. Inspired by this success, in the present work, we focus on the challenging task of making predictions on highly imbalanced data sets. All data sets used for benchmarking the proposed BTDT-MBO method have an IR of 16.5 or greater.  

\vspace{0.5cm}

\begin{table}[H]
    \centering
    \begin{tabular}{c|c|c|c|c}
    Data Set & Data Set Grouping & \# Compounds & Imbalance Ratio (IR) & Labels\\
    \hline
    CHEMBL1909150  & DrugMatrix & 842 & 16.9 & Active, Inactive \\
    CHEMBL1909157   & DrugMatrix & 842 & 18.6 & Active, Inactive \\
    CHEMBL1909132  & DrugMatrix & 842 & 20.0 & Active, Inactive \\
    CHEMBL1909134 & DrugMatrix & 842 & 16.5 & Active, Inactive\\
    CHEMBL100  & DS1 & 2100 & 20.0 & Active, Inactive \\
    CHEMBL8  & DS1 & 2100 & 20.0 & Active, Inactive \\
    \end{tabular}
\caption{Information about the data used for benchmarking the model (results are shown in Figures \ref{dm-fig} and \ref{ds1-fig}). All data sets group molecules into two classes. The first four data sets are from \cite{esposito2021, DrugMatrix} and the last two data sets are from \cite{esposito2021, riniker2013, riniker20132}.}
\label{data-tab}
\end{table}

\subsection{Evaluation Metrics}\label{metrics}

There are various existing evaluation metrics for data classification tasks. One such metric is model accuracy, which computes the fraction of data points correctly classified by the model. However, this metric does not capture many nuances of model performance. Particularly when dealing with imbalanced data, as Esposito et al. discussed in \cite{esposito2021}, the minority class may be the class of most interest, but points from the minority class are more likely to be misclassified by the model than points from the majority class(es). In highly imbalanced data sets, such as the ones used in the present work, there are comparatively very few points in the minority class. Even if the model misclassifies all of the points in the minority class, the performance of the technique could still be very high, and the model could mistakenly appear to be suitable for the highly imbalanced classification task. Thus, to better quantify the performance of our proposed method, we use another evaluation metric, which computes the area under the receiver operating characteristic curve (the ROC curve). This is also known as the ROC-AUC score.\\

\begin{figure}
    \centering
    \begin{minipage}{0.5\textwidth}
        \centering
        \includegraphics[width=\textwidth]{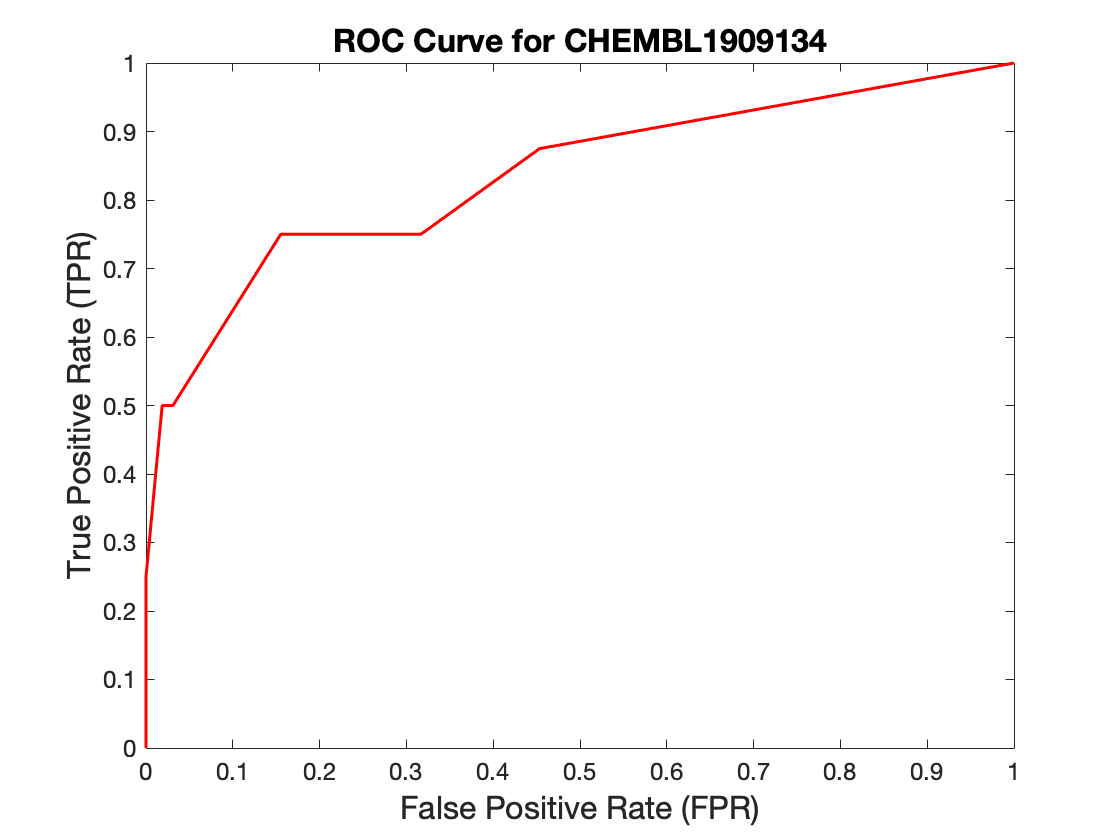}
    \end{minipage}\hfill
    \begin{minipage}{0.5\textwidth}
        \centering
        \includegraphics[width=\textwidth]{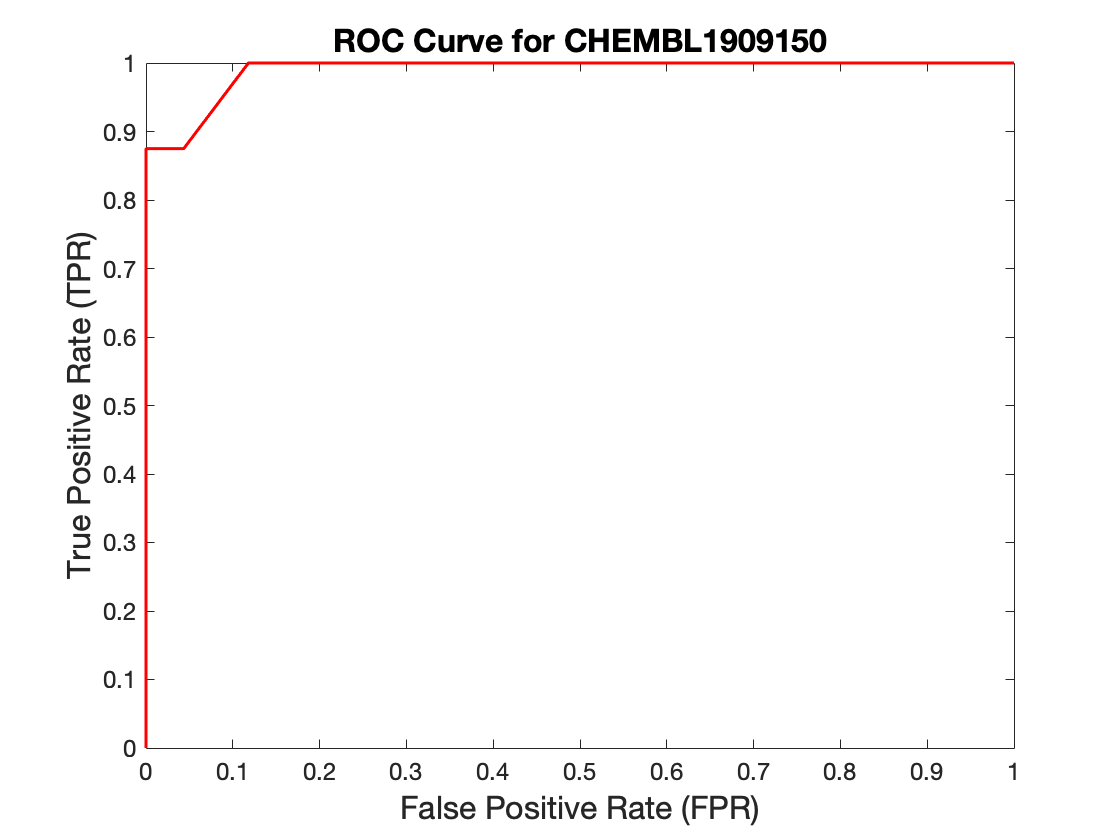}
    \end{minipage}
    \caption{Example ROC curves for two of the data sets (CHEMBL1909134 and CHEMBL1909150) used for benchmarking the proposed method. Curves were constructed in MATLAB for a random labeled/unlabeled split of each data set using a set of increasing thresholds from 0 to 1, with each successive threshold increasing by 0.05. The average ROC-AUC scores for both data sets over all 50 random partitions can be seen in Figure \ref{dm-fig}.}
    \label{roc-fig}
\end{figure}

For the data classification problem, the ROC curve plots the true positive rate versus the false positive rate of the classifier for various thresholds. The ROC curve captures useful information about the classifier, particularly on highly imbalanced data--because the minority (or positive, in our case) class can be easily misclassified, both the true positive rate (TPR) and the false positive rate (FPR) need to be considered. Recall that a classification threshold is used to predict a given data point's class using its probability distribution over all classes. If the probability that the data point is affiliated with the minority class is greater than the threshold, then that point is allotted to the minority class. For a threshold equal to 0, the model would predict that every data point belonged to the minority class. In this case, the model would successfully classify all of the true positive data points (i.e., a TPR of 1), but it would also incorrectly classify all of the negative data points as positive (i.e., an FPR of 1). As the threshold increased, fewer data points would be classified as positive. For a threshold equal to 1, the model would classify all data points as negative, producing a TPR and an FPR of 0.\\

The ROC-AUC score quantifies the relationship between the TPR and FPR by computing the area under the ROC curve. For an area close to 1, it is theoretically possible to choose an "optimal" threshold that yields a high TPR and a low FPR (i.e., produces a point close to the upper-left corner of the ROC plot). In general, an AUC of 0.5 suggests that that algorithm has no ability to distinguish between the classes, 0.7 to 0.8 indicates an acceptable technique, 0.8 to 0.9 describes an excellent technique, and greater than 0.9 designates an outstanding technique. To calculate the ROC-AUC score in our experiments for the proposed BTDT-MBO method, we used the \texttt{perfcurve} function in MATLAB. There also exist functions that calculate the ROC-AUC score for the multiclass case, so our evaluation procedure can be applied for any number of classes.\\

To further illustrate the construction of the ROC curves as well as their reflection of the classification performance of models, we have included example ROC curves in Figure \ref{roc-fig} for two data sets used in benchmarking the proposed BTDT-MBO model. These plots were generated using a random labeled/unlabeled split of each data set, and the thresholds used to construct each plot ranged from 0 to 1, with each threshold increasing by 0.05. The ROC curve on the right for the CHEMBL1909150 data set demonstrates near-ideal performance on this data partition, with a point very close to the upper-left corner of the plot, corresponding to an FPR close to 0 and a TPR close to 1. Indeed, as displayed in Figure \ref{dm-fig} and discussed in Section \ref{performance-and-discussion}, our model's average ROC-AUC score for 50 random partitions of the CHEMBL1909150 data set was 0.978, displaying outstanding classification performance. The ROC curve on the left for the CHEMBL1909134 data set clearly has a lower ROC-AUC value than the curve for the CHEMBL1909150 data set, but the model still shows some ability to discriminate between the two classes. As shown in Figure \ref{dm-fig}, our model's average ROC-AUC score over all partitions on the CHEMBL1909134 data set was 0.792, which is not surprising given its example ROC curve. Do note, however, that our proposed model performed superior to the comparison methods on this data set, which is further discussed in Section \ref{performance-and-discussion}. \\

\subsection{Model Performance and Discussion}\label{performance-and-discussion}

To evaluate the performance of our BTDT-MBO method for highly imbalanced molecular data sets, we compare our models against the GHOST algorithm from Esposito et al. \cite{esposito2021}. The GHOST algorithm is designed to be paired with any machine learning classifier, and the code used in \cite{esposito2021} ran experiments using GHOST with random forest (RF), extreme gradient boosting (XGB), logistic regression (LR) and gradient boosting (GB) classifiers. In the present work, all models used 80\% of the data as training (or labeled, in the MBO case) to mirror the experiments by Esposito et al. \cite{esposito2021}.\\

\begin{figure}
    \centering
    \begin{minipage}{0.5\textwidth}
        \centering
        \includegraphics[width=\textwidth]{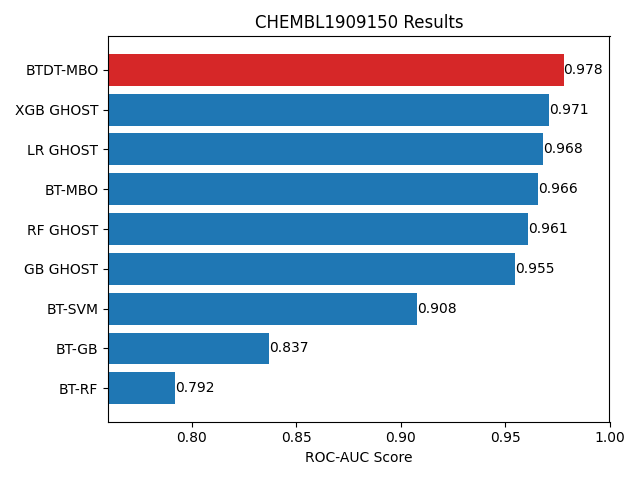}
    \end{minipage}\hfill
    \begin{minipage}{0.5\textwidth}
        \centering
        \includegraphics[width=\textwidth]{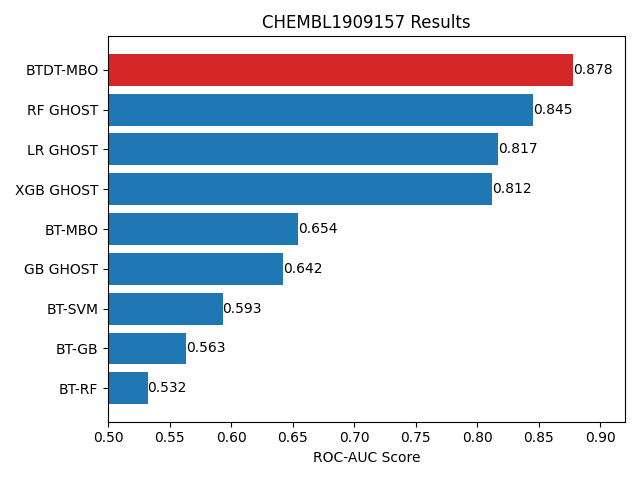}
    \end{minipage}
    \begin{minipage}{0.5\textwidth}
        \centering
        \includegraphics[width=\textwidth]{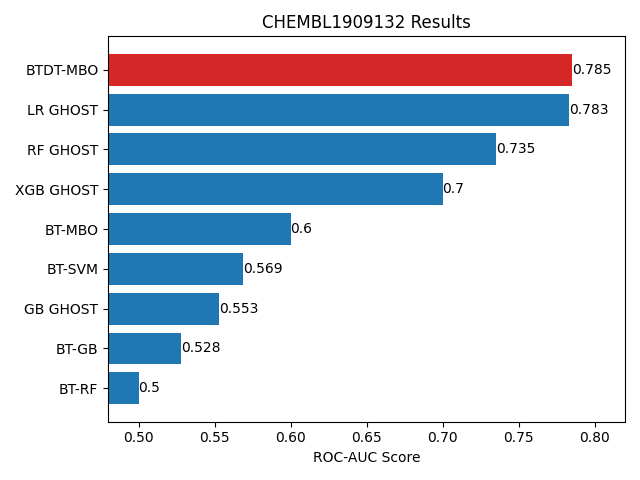}
    \end{minipage}\hfill
    \begin{minipage}{0.5\textwidth}
        \centering
        \includegraphics[width=\textwidth]{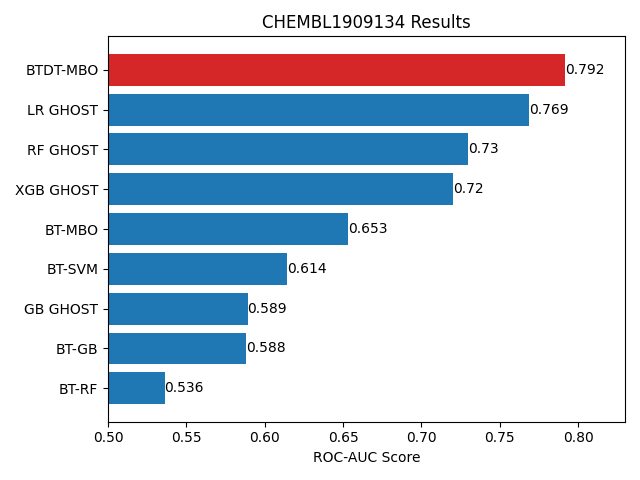}
    \end{minipage}
    \caption{Comparison to other techniques on DrugMatrix data sets. The results of our proposed method are in red, while those of other algorithms are in blue. The imbalance ratios for the pictured data sets vary from 16.5 to 20.0. Detailed information about the overall size and composition of the comparison data sets is described in Section \ref{data-sets}. The performance metric is the ROC-AUC score averaged over 50 random training-testing (or labeled/unlabeled) splits of the given data, with 80\% of the data being labeled in each case. The BTDT-MBO result for each data set is the highest of the BTDT-MBO model using a Gaussian weight function and the BTDT-MBO model using a distance correlation weight function. Some comparison results were generated using the GHOST algorithm \cite{esposito2021} via random forests (RF), extreme gradient boosting (XGB), logistic regression (LR) and gradient boosting (GB). Additional comparison results were generated using the BT-MBO algorithm \cite{hayes2023} as well as BT-GB, BT-RF, and BT-SVM models (consisting of the BT-FPs passed to gradient boosting, random forest, and support vector machine algorithms, respectively).}
    \label{dm-fig}
\end{figure}

\begin{figure}
    \centering
    \begin{minipage}{0.5\textwidth}
        \centering
        \includegraphics[width=\textwidth]{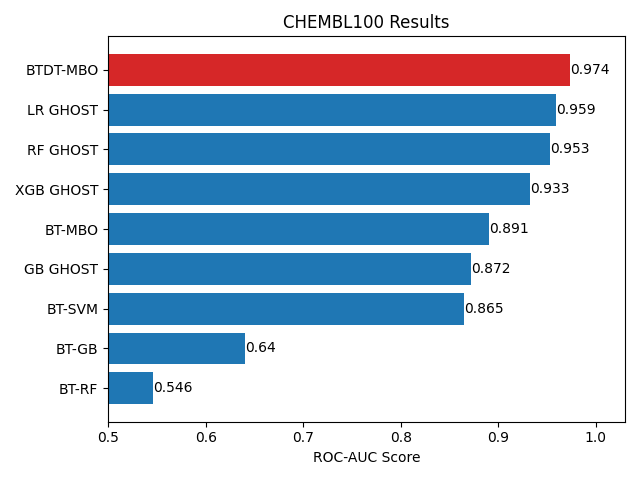}
    \end{minipage}\hfill
    \begin{minipage}{0.5\textwidth}
        \centering
        \includegraphics[width=\textwidth]{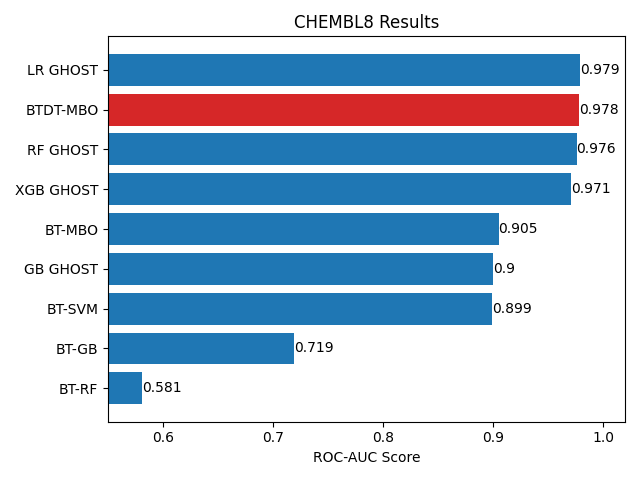}
    \end{minipage}
    \caption{Comparison to other techniques on two DS1 data sets. The results of our proposed method are in red, while those of other algorithms are in blue. The imbalance ratios for the pictured data sets are both 20.0. Detailed information about the overall size and composition of the comparison data sets is described in Section \ref{data-sets}. The metric is the ROC-AUC score averaged over 50 splits of the data, with 80\% of the data being labeled in each case. The BTDT-MBO result for each data set is the highest of the BTDT-MBO model using a Gaussian weight function and the BTDT-MBO model using a distance correlation weight function. Some comparison results were generated using the GHOST algorithm \cite{esposito2021} via random forests (RF), extreme gradient boosting (XGB), logistic regression (LR) and gradient boosting (GB). Additional comparison results were generated using the BT-MBO algorithm \cite{hayes2023} as well as BT-GB, BT-RF, and BT-SVM models (consisting of the BT-FPs passed to gradient boosting, random forest, and support vector machine algorithms, respectively).}
    \label{ds1-fig}
\end{figure}
Figures \ref{dm-fig} and \ref{ds1-fig} display the results of the BTDT-MBO procedure compared with the GHOST method from Esposito et al. \cite{esposito2021}. We generated results using GHOST paired with LR, RF, GB, and XGB. The GHOST model allows the user to select the desired classification metric to use for optimization from two options: the Cohen's kappa, which is a balanced classification metric that was used as the default metric in \cite{esposito2021}, or the ROC-AUC score. All of the experiments in this paper were evaluated using the ROC-AUC score.\\

Overall, the experimental results indicate that the BTDT-MBO algorithm obtains higher ROC-AUC scores than the comparison methods for all data, except CHEMBL8, for which the result of the proposed algorithm is second best, but almost the same as the best. We first outline the details of the experiments in this work, after which we provide a summary as well as a further discussion of results.\\

To create labeled and unlabeled partitions in the data input to our DT-MBO method, we randomly selected 80\% of the points to be labeled and denoted the rest as unlabeled. Then, we applied the DT-MBO method, which propagates the known labels to the unlabeled nodes for a prescribed number of iterations as detailed in Section \ref{mbo}. We used the resulting predicted labels to calculate the optimal ROC-AUC score for this partition (over all thresholds). In highly imbalanced data sets, which often contain very few points in the minority class, results can vary widely depending on the particular training/testing or labeled/unlabeled partition. To account for this variability, as in \cite{esposito2021}, we repeated each experiment 50 times, with a new random partition for each experiment, and averaged the ROC-AUC scores over the 50 trials. We used a similar splitting and averaging procedure to generate the GHOST comparison results; however, Esposito et al. \cite{esposito2021} recorded the random seeds used for each training/testing data split in their experiments, and we used the same 50 random seeds when running the GHOST tests. Results in Figures \ref{dm-fig} and \ref{ds1-fig} indicate the average ROC-AUC score over 50 different splits of the data.\\

Regarding parameters, the proposed BTDT-MBO model requires specification of various parameters, whose functions can be seen in Algorithms \ref{alg-bt} and \ref{alg-dtmbo}. The bidirectional transformer \cite{hundreds} requires the user to choose one of three pretraining data sets as outlined in Section \ref{bt}. In the present work, all BTDT-MBO experiments use the transformer model trained only on the ChEMBL data set \cite{chembl-source}, based on its success in predicting scarcely labeled molecular data in experiments from the authors' previous work \cite{hayes2023} using the BT-MBO method. The transformer model does not require any additional parameter specifications, as all other parameters are implicit in the pretrained models \cite{hundreds, gao2021proteome}.\\

The DT-MBO algorithm relies on several parameters, many of which are discussed in our previous work \cite{hayes2023}, but some of which have been added in the present work to implement the decision threshold adjustment. Some parameters were defined manually prior to running all experiments, and other parameters were tuned in our experiments. Here, we provide a brief description of all parameters and their settings or tuning. First, to create the framework in which the DT-MBO algorithm operates, we construct a $N_n$-neighbor graph. After, the graph Laplacian $\mathbf{L}$ (\ref{laplacian}) is constructed, and the first (smallest) $N_e$ eigenvalues and corresponding eigenvectors of $\mathbf{L}$ are computed. Overall, both $N_n$ and $N_e$ were designated as hyperparameters and tuned in our experiments. Other hyperparameters for the DT-MBO method that were tuned are the $C$ and $dt$ parameters in the diffusion step, as well as the number of iterations, $N_t$.  

Regarding other parameters, we let $N_s=3$ and $N_l=50$. Additionally, the number of labeled points $N_p$ was set at 80\% of each data set for all tasks. Finally, the decision threshold adjustment steps in our algorithm require the specification of a minimum threshold and a maximum threshold to test. For all BTDT-MBO experiments, we used a minimum threshold of 0.05 and a maximum threshold of 0.55.\\

As previously stated, the proposed BTDT-MBO algorithm obtained higher ROC-AUC scores than the compared methods in all but one of the six highly imbalanced data sets included in this work. Figure \ref{dm-fig} displays our model's results (using the best-performing of the BTDT-MBO with Gaussian weight function and BTDT-MBO with distance correlation weight function) compared with the GHOST algorithm's results on four highly imbalanced DrugMatrix data sets. The BTDT-MBO procedure obtained a higher ROC-AUC score than all GHOST models for all four of the DrugMatrix data sets. As indicated in Table \ref{data-tab}, each of the four DrugMatrix data sets each contained 842 compounds and had IRs of at least 16.5. Our model achieved its best predictive performance on the CHEMBL1909150 data set with an average ROC-AUC score of 0.978, indicating an outstanding ability to discriminate between the active and inactive classes. Furthermore, our model outperformed all GHOST models, which also demonstrated very high ROC-AUC scores on this data set. For CHEMBL1909157, our procedure obtained excellent results, with an ROC-AUC score of 0.866. Some of the GHOST models also achieved ROC-AUC scores greater than 0.8, but the GHOST model using GB performed significantly worse. Our model similarly outperformed all four GHOST comparison models on the CHEMBL1909132 and CHEMBL1909134 data sets.\\

Figure \ref{ds1-fig} shows our model's results on two highly imbalanced DS1 data sets, again compared with the GHOST algorithm paired with LR, RF, XGB, and GB. Recall from Section \ref{data-sets} that all DS1 data sets have an IR of 20.0 by construction. For CHEMBL100, our method achieved an outstanding ROC-AUC score of 0.974, while the highest-performing GHOST model (using LR) earned an ROC-AUC score of 0.959. While our BTDT-MBO method did not achieve the highest ROC-AUC score of comparison algorithms on CHEMBL8, it obtained the second-best ROC-AUC score, performing almost as well as the top-performing GHOST model using LR. All models in this case demonstrated ROC-AUC scores of at least 0.9, with the highest method earning a score of 0.979 and BTDT-MBO scoring 0.978.\\

As outlined in Section \ref{distcorr}, in addition to the default Gaussian kernel used in our BTDT-MBO method, we also conducted experiments that instead used the distance correlation to compute the weights between data elements. In our experiments, the BTDO-MBO method using a Gaussian weight function yielded higher ROC-AUC scores than the BTDT-MBO method using distance correlation for five out of the six data sets included in this work, with the BTDT-MBO algorithm using distance correlation achieving the best results on the CHEMBL1909157 DrugMatrix data set. However, the BTDT-MBO algorithm using a distance correlation weight function performed similarly to (and frequently almost as well as) the BTDT-MBO method using a Gaussian kernel on all six data sets, particularly on the DS1 data sets used for benchmarking. For the CHEMBL100 DS1 data set, the BTDT-MBO model using distance correlation yielded an ROC-AUC score of 0.970, slightly lower than BTDT-MBO using Gaussian weights (ROC-AUC = 0.974) but still higher than all comparison methods. Similarly, for the CHEMBL8 DS1 data set, BTDT-MBO using distance correlation scored 0.973, which was again lower than BTDT-MBO using a Gaussian weight function (ROC-AUC = 0.978) but higher than two comparison methods. All tested models scored very highly on this data set.\\

Our model using distance correlation saw similar results on the DrugMatrix data sets. In fact, as mentioned above, BTDT-MBO using distance correlation achieved the best overall result of all tested models on the CHEMBL1909157 DrugMatrix data set, earning an ROC-AUC score of 0.878. BTDT-MBO using a Gaussian kernel scored an ROC-AUC value of 0.866; thus, both BTDT-MBO methods beat all tested GHOST models. For the CHEMBL1909150 data set, BTDT-MBO using distance correlation scored an ROC-AUC value of 0.975, higher than all comparison methods and only lower than our BTDT-MBO model using a Gaussian, which scored 0.978. On the CHEMBL1909132 DrugMatrix data set, BTDT-MBO using distance correlation earned an ROC-AUC score of 0.743, which places it below BTDT-MBO using Gaussian weights (ROC-AUC = 0.785) and GHOST using LR, but above the three other GHOST comparison methods. Our model using distance correlation saw similar relative results for the CHEMBL1909134 DrugMatrix data set, with an average ROC-AUC score of 0.766. The model earned the third-highest score of all methods tested for this data set, lower than our BTDT-MBO method using Gaussian weights (ROC-AUC = 0.792) and very close to GHOST paired with LR, which scored a 0.769. Overall, these results suggest that the distance correlation can be useful for quantifying similarity in molecular data sets, including those that are highly imbalanced. Furthermore, the distance correlation may be considered as a potential alternative weight function to the Gaussian function in similarity graph-based settings, particularly when the data is not close to a normal distribution.\\

\begin{figure}
    \centering
    \includegraphics[width=0.6\textwidth]{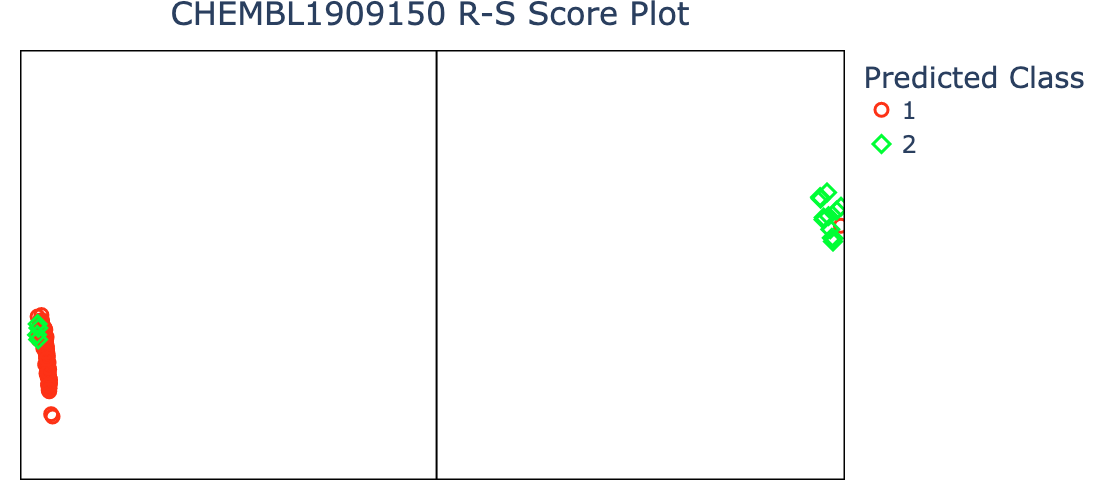}\vspace{6pt}
    \includegraphics[width=0.6\textwidth]{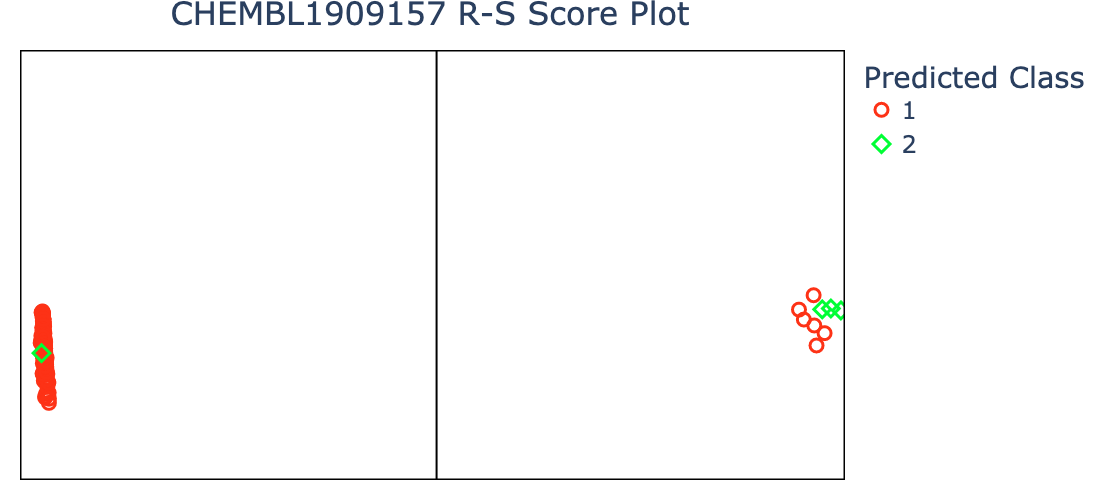}\vspace{6pt}
    \includegraphics[width=0.6\textwidth]{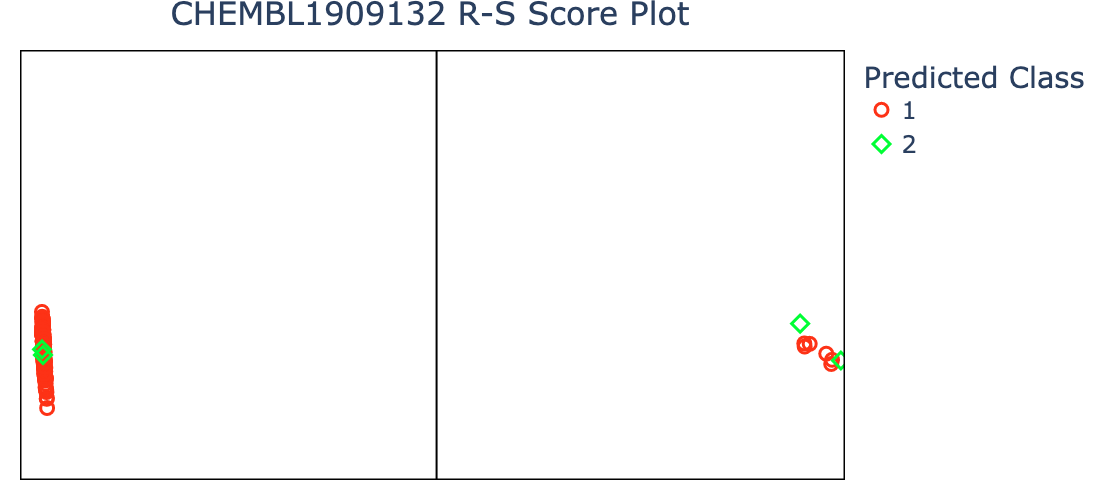}\vspace{6pt}
    \includegraphics[width=0.6\textwidth]{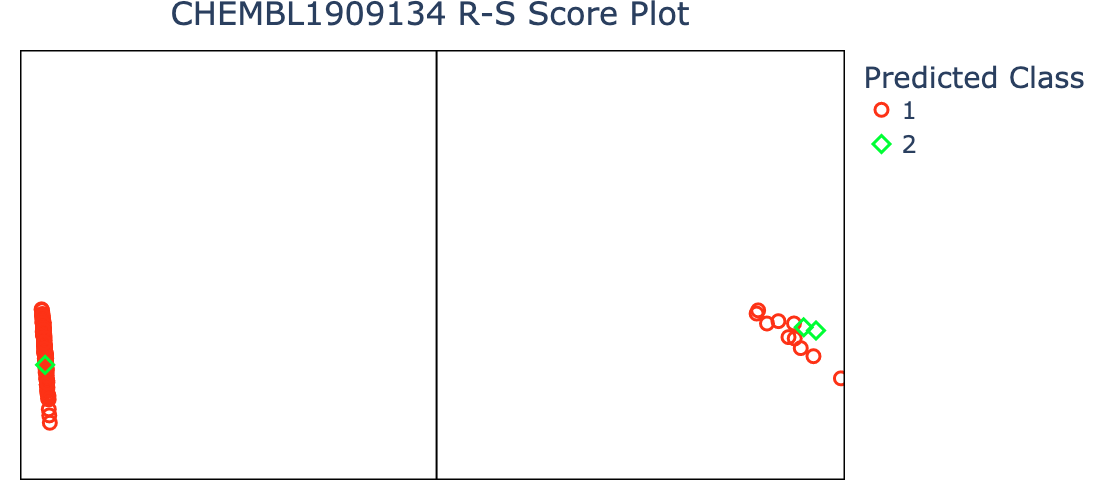}
    \caption{R-S score plots for the four DrugMatrix data sets. The plots display R-S scores of unlabeled points from a random labeled/unlabeled partition of each data set. The $x$ and $y$ axes of the plots represent residue and similarity scores, respectively. From left to right, the panels plot points in class 1 (i.e., inactive compounds) and class 2 (i.e., active compounds). Each point is colored based on its class predicted by the proposed BTDT-MBO model.}
    \label{dm-rs-fig}
\end{figure}

\begin{figure}
    \centering
    \includegraphics[width=0.53\textwidth]{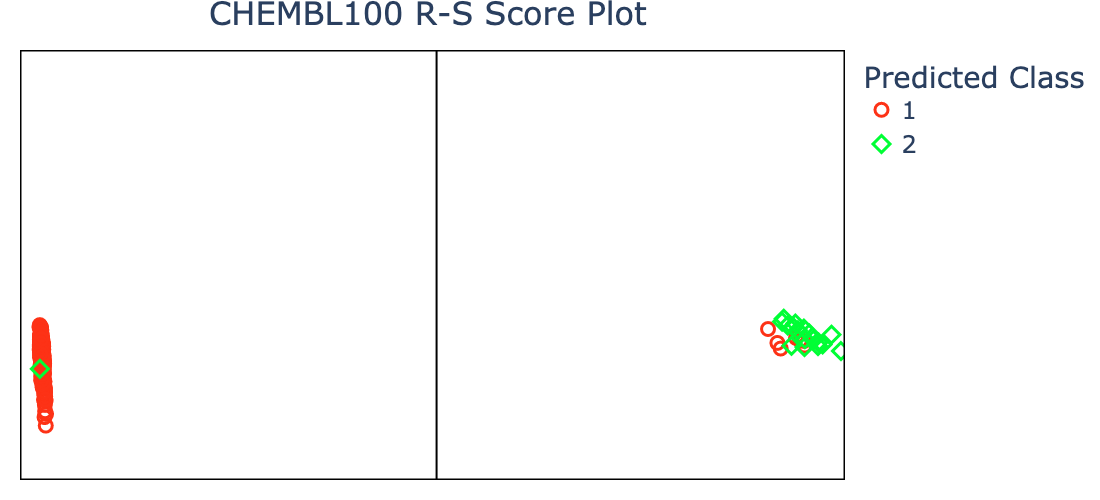}\vspace{3pt}
    \includegraphics[width=0.53\textwidth]{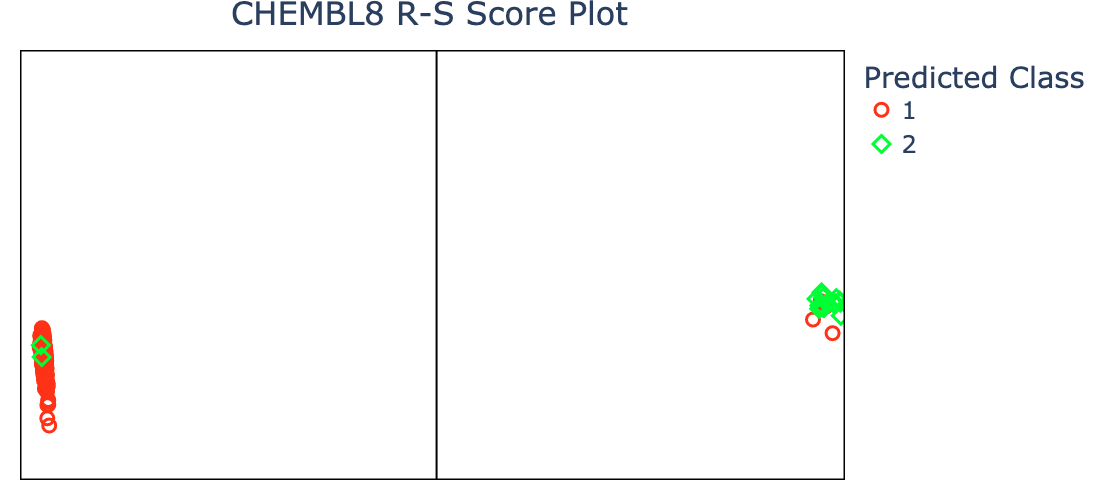}
    \caption{R-S score plots for the two DS1 data sets. The plots display R-S scores of unlabeled points from a random labeled/unlabeled partition of each data set. The $x$ and $y$ axes of the plots represent residue and similarity scores, respectively. From left to right, the panels plot points in class 1 (i.e., inactive compounds) and class 2 (i.e., active compounds). Each point is colored based on its class predicted by the proposed BTDT-MBO model.}
    \label{ds1-rs-fig}
\end{figure}

To further visualize our proposed model's performance on the benchmark data sets, we additionally utilized the residue-similarity (R-S) scores and associated R-S plots introduced by Hozumi et al. \cite{ccp}. For a given data point in a particular class, its residue score is defined as the sum of the distances from that point to points in other classes. The point's similarity score is defined as the average distance from that point to points in its own class. In other words, the residue score measures how well data points in a given class are separated from other classes, and the similarity score measures how well data points in a given class are clustered together. Given a training/testing (or labeled/unlabeled) split of a data set, one can construct R-S score plots visualizing the elements of each class separately. In Figures \ref{dm-rs-fig} and \ref{ds1-rs-fig}, we plot R-S scores for a random labeled/unlabeled split of the six data sets used in our work. Points are separated into panels by their true class label, and the color of each point corresponds to its predicted class using our proposed BTDT-MBO model. We use the optimal weight function for each data set as discussed above.\\

Another novel addition to the proposed BTDT-MBO method is its threshold optimization step, the details of which are given in Section \ref{mbo}. This step chose the decision threshold that yielded the highest ROC-AUC score for a given partition of a particular data set. Because the partitions were created randomly, the class distribution of the unlabeled and labeled portions of a data set could vary. Additionally, due to the high level of class imbalance in the data, there may be very few data points from the minority class in the labeled or unlabeled set for a particular partition. Therefore, it is not surprising that in our experiments, the optimal thresholds for the splits varied considerably. However, the optimal thresholds often tended to be lower than the typical threshold of 0.5. For example, in one experiment on the CHEMBL1909157 data set (i.e., testing over 50 random labeled/unlabeled partitions), the lowest optimal threshold for the 50 random splits was 0.05, and the highest optimal threshold was 0.45. The median optimal threshold over all 50 splits was 0.15. These results reinforce the impact and the advantage of adjusting the decision threshold on classifier performance for highly imbalanced data. 

\subsection{Statistical Testing}

In this paper, we perform statistical analysis using the experimental results; we perform the Friedman and the post-hoc Nemenyi test in order to achieve a more thorough evaluation. The results are shown in Table 6 and Figure 7, including the mean rank values, where smaller ranks demonstrate more competitive techniques. The critical distance was around 4.29, with the significance level being 0.05. Overall, the Friedman test's null hypothesis is that all algorithms have the same performance. If the p-value is less that 0.05, the null hypothesis is rejected, which it was in our testing. Moreover, the Nemenyi post-hoc test is utilized to identify which procedures are significantly different from each other. Overall, it is shown that the proposed algorithm is ranked consistently the best among comparison methods. 

\subsection{Future Work and Limitations}

While the proposed method has many advantages, such as the ability to perform accurately with highly imbalanced data, it has certain limitations similarly to any other algorithm. For example, the proposed procedure requires the calculation of a small number of the graph Laplacian's eigenvectors and corresponding eigenvalues. This can obviously be computationally expensive for certain data sets. One possible direction in such a scenario is to construct an approximation of the full graph using other techniques, e.g., sampling-based procedures, in particular the Nystr\"{o}m Extension technique, detailed in sources such as \cite{nystrom1,nystrom2,nystrom3}. Such a technique can be utilized to very efficiently compute approximations of the eigenvalues and eigenvectors for the proposed procedure even for large data sets.\\

Future work regarding this study includes exploring imposing class size constraints into the technique. Moreover, we would like to explore integrating other correlation techniques, as well as incorporating extended-connectivity fingerprints or autoencoders into the algorithm structure.

\begin{table*}[h!]
\vspace{0.5cm}
     \renewcommand\thetable{6}
\caption{\hspace{0.2cm} Results for Statistical Testing}
\vspace{0.2cm}
\label{table:friedman}
 \begin{subtable}{0.5\textwidth}
  \caption{\hspace{0.2cm} Reference for Tables 5b-5c}
  \vspace{0.2cm}
    \centering
        \begin{tabular}{| c || c || c | c || c | c | }
\hline
 &  Method & &  Method & &  Method \\
 \hline
`1'  &  BTDT-MBO (Proposed) & `4' & XGB-GHOST  & `7'   & BT-SVM   \\
`2'  & LR-GHOST  &  `5'  &  BT-MBO  & `8' &   BT-GB   \\
`3'  &  RF-GHOST & `6' &  GB-GHOST  & `9'   &   BT-RF \\
 \hline
        \end{tabular}
\end{subtable}
\hspace{0.35cm}
 \begin{subtable}{0.6\textwidth}
   \caption{\hspace{0.2cm} Results of Friedman's Test}
   \vspace{0.2cm}
       \centering
        \begin{tabular}{| c ||  c | c |  }
 \hline
   Chi-sq statistic &  p-value \\
 \hline
   112.39130  &   2.4237e-07 \\

 \hline
        \end{tabular}
\end{subtable}%
\vspace{0.3cm}
 \begin{subtable}{0.8\textwidth}
\hspace{0.2cm}
\vspace{0.225cm}
\hspace{0.75cm}
\parbox{.8\linewidth}{
\vspace{0.5cm}
\caption{\hspace{0.2cm} Mean rank values of the Friedman/Nemenyi test}
\vspace{0.2cm}
\scalebox{.975}{
\begin{tabular}{| c | c | c | c | c | c | c | c | c | c |}
\hline
`1' & `2' & `3' & `4' & `5' & `6' & `7' & `8' & `9' \\
 \hline
 1.16667 & 2.16667 & 3,16667 & 3.66667 & 4.83333 & 6.33333 & 6.66667 & 8.00000 & 9.00000 \\
 \hline
\end{tabular}}
}
\end{subtable}
\end{table*}

\vspace{-0.2cm}
 
\begin{figure*}[h!]
    \centering
    \hspace{-2.5cm}
    \begin{minipage}{0.8\textwidth}
        \centering
        \includegraphics[width=\textwidth]{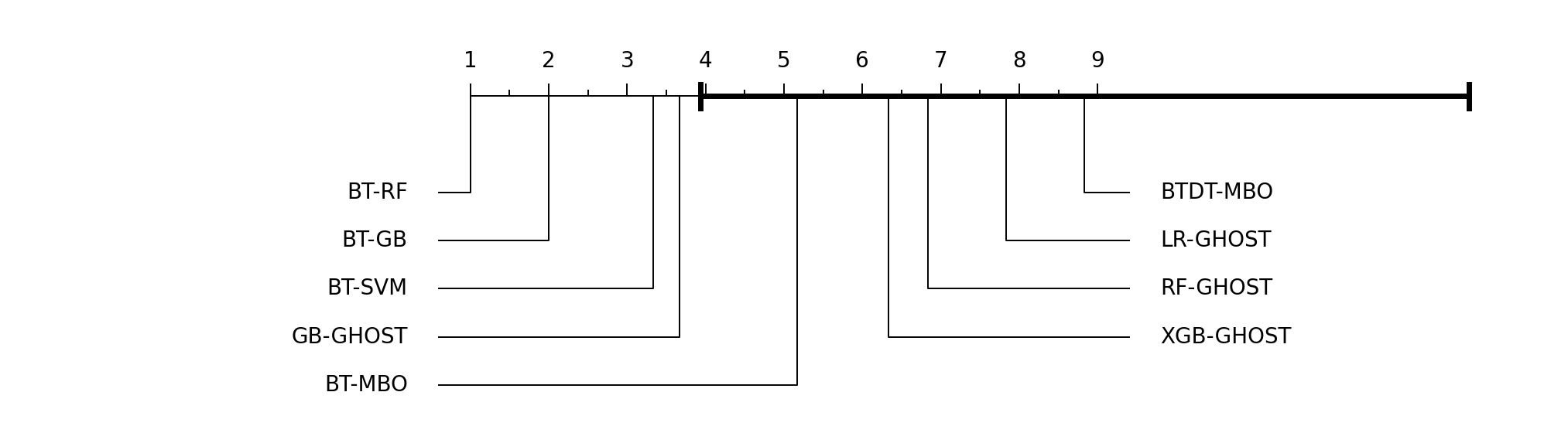}
    \end{minipage}\hfill
    \caption{The average rank diagram of the Friedman/Nemenyi test}
    \label{ds1-fig}
\end{figure*}

\section{Conclusion}

Given the prevalence of molecular data where the class sizes vary vastly, the ability of an algorithm to accurately predict the class of data elements in imbalanced data sets is extremely important. However, many classification techniques are not accurate on highly imbalanced data as they tend to overestimate majority classes. This paper presents a BTDT-MBO method for molecular data classification in case of imbalanced data sets with highly varied class sizes.  The method, which integrates Merriman-Bence-Osher procedures as well as a bidirectional transformer, incorporates adjustments in the data classification threshold for machine learning classifiers to handle the high class imbalance. To illustrate the advantages of the proposed technique, particular attention is given to highly imbalanced data sets. \\

Numerical experiments indicate that the BTDT-MBO procedure performs very well even with high class imbalance ratios; in particular, the computational experiments on DS1 data sets and DrugMatrix data sets demonstrate that the results of the proposed procedure are almost always more accurate than the comparison algorithms such as \cite{esposito2021}, as indicated by Figures 2 and 3. The ROC-AUC score was used as a metric. Overall, the new model serves as a powerful machine learning tool for data sets with a high class imbalance.\\

The present work additionally investigates the distance correlation as a choice for a weight function in the proposed BTDT-MBO algorithm, serving as an alternative to the Gaussian kernel often used to compute graph weights. Overall, the results for the BTDT-MBO technique using a distance correlation weight function are similar to those using Gaussian weights, with the distance correlation model performing better for one data set; these results suggest further utility of the distance correlation in imbalanced molecular data classification and similarity graph-based methods. In particular, data that is far from a normal distribution may benefit from the use of a distance correlation weight function rather than a Gaussian weight function.\\

\section*{Funding Information}
The work of this paper is supported partly by the NSF DMS-2052983 grant. The third author was supported partly by NIH R01GM126189, R01AI164266 and R35GM148196 grants, NSF grants DMS-1761320, DMS-2052983, and IIS-1900473,  NASA grant 80NSSC21M0023,  Michigan Economic Development Corporation, MSU Foundation, Pfizer, and Bristol-Myers Squibb 65109.  

\section*{Conflict of Interest}
All authors declare that they have no conflicts of interest.

\section*{Author Contributions}
Nicole Hayes was responsible for the software, computational experiments and writing. Ekaterina Rapinchuk and Guo-Wei Wei were responsible for methodology, reviewing, editing, some writing and supervision.

\section*{Data Availability}
The data sets in this paper are publically available at \url{https://github.com/rinikerlab/GHOST/tree/main/data}.

\section*{Statement of Usage of Artificial Intelligence}
The authors have not received any help from artificial intelligence such as ChatGPT while preparing this paper.

\bibstyle{achemso}
\bibliography{bib_draft.bib}


\end{document}